\documentclass[letterpaper]{article} 
\usepackage{aaai23}  
\usepackage{times}  
\usepackage{helvet}  
\usepackage{courier}  
\usepackage[hyphens]{url}  
\usepackage{graphicx} 
\usepackage{natbib}  
\usepackage{caption} 
\usepackage{amssymb}
\usepackage{bm}
\usepackage[ruled,linesnumbered,noend]{algorithm2e}
\usepackage{tikz}
\usepackage{comment}
\usepackage{amsmath} 
\usepackage{subcaption}
\usepackage{bbding}
\usepackage{booktabs}
\usepackage{enumitem}
\usepackage{multirow}
\usepackage{bm}
\usepackage{colortbl}
\usepackage{url}
\usepackage{bbding}

\title{Combating Unknown Bias with \\Effective Bias-Conflicting Scoring and Gradient Alignment}
\author {
    Bowen Zhao\textsuperscript{\rm 1},
    Chen Chen\textsuperscript{\rm 2,\Envelope},
    Qian-Wei Wang\textsuperscript{\rm 1,\rm 3},
    Anfeng He\textsuperscript{\rm 2},
    Shu-Tao Xia\textsuperscript{\rm 1,\rm 3,\Envelope}
}
\affiliations {
    \textsuperscript{\rm 1}Tsinghua University,
    \textsuperscript{\rm 2}Tencent TEG AI,
    \textsuperscript{\rm 3}Peng Cheng Laboratory\\
    zbw18@mails.tsinghua.edu.cn, chen1634chen@gmail.com, xiast@sz.tsinghua.edu.cn
}

\begin{document}

\maketitle

\begin{abstract}
Models notoriously suffer from dataset biases which are detrimental to robustness and generalization. The identify-emphasize paradigm shows a promising effect in dealing with unknown biases. However, we find that it is still plagued by two challenges: A, the quality of the identified bias-conflicting samples is far from satisfactory; B, the emphasizing strategies just yield suboptimal performance. In this work, for challenge A, we propose an effective bias-conflicting scoring method to boost the identification accuracy with two practical strategies --- peer-picking and epoch-ensemble. For challenge B, we point out that the gradient contribution statistics can be a reliable indicator to inspect whether the optimization is dominated by bias-aligned samples. Then, we propose gradient alignment, which employs gradient statistics to balance the contributions of the mined bias-aligned and bias-conflicting samples dynamically throughout the learning process, forcing models to leverage intrinsic features to make fair decisions. Experiments are conducted on multiple datasets in various settings, demonstrating that the proposed solution can alleviate the impact of unknown biases and achieve state-of-the-art performance.
\end{abstract}

\section{Introduction}
Deep Neural Networks (DNNs) have achieved significant progress in various visual tasks. DNNs tend to learn \textbf{intended} decision rules to accomplish target tasks commonly. However, they may follow \textbf{unintended} decision rules based on the easy-to-learn shortcuts to ``achieve" target goals in some scenarios~\cite{bahng2020learning}. For instance, when training a model to classify digits on Colored MNIST~\cite{kim2019learning}, in which the images of each class are mainly dyed by one pre-defined color respectively (\textit{e.g.}, most `0' are red, `1' are yellow), the intended decision rules classify images based on the shape of digits, while the unintended decision rules utilize color information instead. Following~\citet{nam2020learning}, sample $x$ that can be ``correctly" classified by unintended decision rules is denoted as a \textbf{bias-aligned} sample $\underline{x}$ (\textit{e.g.}, red `0') and vice versa a \textbf{bias-conflicting} sample $\overline{x}$ (\textit{e.g.}, green `0').

\begin{figure}[htbp]
\centering
\includegraphics[width=0.9\columnwidth]{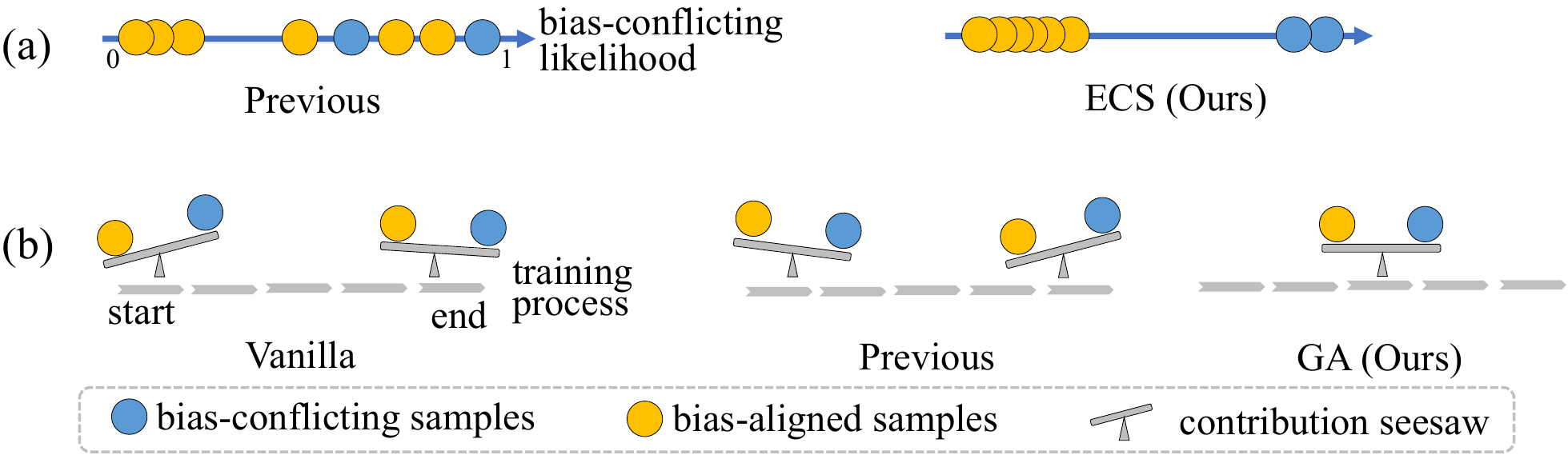}
\caption{(a) Effective bias-Conflicting Scoring helps identify real bias-conflicting samples in stage \uppercase\expandafter{\romannumeral1}. (b) Gradient Alignment balances contributions from the mined bias-aligned and bias-conflicting samples throughout training, enforcing models to focus on intrinsic features in stage \uppercase\expandafter{\romannumeral2}.
}
\label{fig:intro_effect} 
\end{figure}

There are also many similar scenarios in real-world, \textit{e.g.}, an animal-centric image set may be biased by the habitats in backgrounds, and a human-centric set can be biased by gender or racial information. Models blinded by biased datasets usually perform poorly in mismatched distribution (\textit{e.g.}, a red `8' may be incorrectly categorized as `0'). Worse still, models with racial or gender bias, \textit{etc.} can cause severe negative social impacts. Moreover, in most real-world problems, the bias information (both bias type and precise labels of bias attribute) is unknown, making debiasing more challenging. Therefore, combating unknown biases is urgently demanded when deploying AI systems in realistic applications.

One major issue that leads to biased models is that the training objective (\textit{e.g.}, vanilla empirical risk minimization) can be accomplished via only unintended decision rules~\cite{sagawa2020investigation}. Accordingly, some studies~\cite{nam2020learning,kim2021learning} try to identify and emphasize the bias-conflicting samples. Nevertheless, we find that the low identification accuracy and the suboptimal emphasizing strategies impair the debiasing effect. In this work, we build an enhanced two-stage debiasing scheme to combat unknown dataset biases. In stage \uppercase\expandafter{\romannumeral1}, we present an Effective bias-Conflicting Scoring (ECS) function to mine bias-conflicting samples. On top of the off-the-shelf method, we propose a peer-picking mechanism to consciously pursue seriously biased auxiliary models and employ epoch-ensemble to obtain more accurate and stable scores. In stage \uppercase\expandafter{\romannumeral2}, we propose Gradient Alignment (GA), which balances the gradient contributions across the mined bias-aligned and bias-conflicting samples to prevent models from being biased. The gradient information is served as an indicator to down-weight (up-weight) the mined bias-aligned (bias-conflicting) samples in order to achieve dynamic balance throughout optimization. The effects of ECS and GA are illustrated in Fig.~\ref{fig:intro_effect}. Compared to other debiasing techniques, the proposed solution (i) does not rely on comprehensive bias annotations or a pre-defined bias type; (ii) does not require disentangled representations, which may fail in complex scenarios where disentangled features are hard to extract; (iii) does not introduce any data augmentations, avoiding additional training complexity such as in generative models; (iv) does not involve any modification of model architectures, making it easy to be applied to other networks. (v) significantly improves the debiasing performance.

\begin{figure*}[t]
\centering
\includegraphics[width=0.75\textwidth]{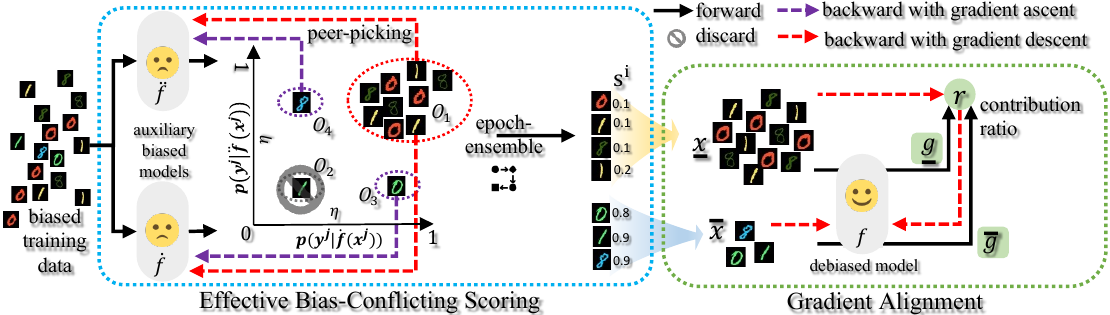}
\caption{Our debiasing scheme. \textbf{Stage \uppercase\expandafter{\romannumeral1}:} training auxiliary biased models $\dot{f}, \ddot{f}$ with peer-picking and epoch-ensemble to score the likelihood that a sample is bias-conflicting (Sec.~\ref{sec:det}). \textbf{Stage \uppercase\expandafter{\romannumeral2}:} learning debiased model $f$ with gradient alignment (Sec.~\ref{sec:ga}). A dashed arrow starting from a sample cluster indicates that the model is updated with gradients from these samples.
}
\label{fig:intro}
\end{figure*}

The main contributions are:
\textbf{(i)} To combat unknown dataset biases, we present an enhanced two-stage approach (illustrated in Fig.~\ref{fig:intro}). We propose an effective bias-conflicting scoring algorithm equipped with peer-picking and epoch-ensemble in stage \uppercase\expandafter{\romannumeral1} (Sec.~\ref{sec:det}), and gradient alignment in stage \uppercase\expandafter{\romannumeral2} (Sec.~\ref{sec:ga}).
\textbf{(ii)} Broad experiments on commonly used datasets are conducted to compare several debiasing methods in a fair manner (overall, we train more than 650 models), among which the proposed method achieves state-of-the-art performance (Sec.~\ref{sec:settings} and Sec.~\ref{sec:quan_com}).
\textbf{(iii)} We undertake comprehensive analysis of the efficacy of each component, hyper-parameters, \textit{etc.} (Sec.~\ref{sec:further_ana}).

\section{Related work}
\label{sec:related}
\noindent \textbf{Combating biases with known types and labels.} Many debiasing approaches require explicit bias types and bias labels for each training sample. A large group of strategies aims at disentangling spurious and intrinsic features~\cite{moyer2018invariant}. For example, EnD~\cite{tartaglione2021end} designs regularizers to disentangle representations with the same bias label and entangle features with the same target label; BiasCon~\cite{hong2021unbiased} pulls samples with the same target label but different bias labels closer in the feature space based on contrastive learning; and some other studies learn disentangled representation by mutual information minimization~\cite{zhu2021learning,kim2019learning,ragonesi2021learning}. Another classic approach is to reweigh/resample training samples based on sample number or loss of different explicit groups~\cite{li2018resound,sagawa2020investigation,li2019repair}, or even to synthesize samples~\cite{agarwal2020towards}. Besides, \citet{Sagawa*2020Distributionally} and \citet{goel2021model} intend to improve the worst-group performance through group distributionally robust optimization~\cite{goh2010distributionally} and Cycle-GAN~\cite{zhu2017unpaired} based data augmentation, respectively. Furthermore, IRM~\cite{arjovsky2019invariant} is designed to learn a representation that performs well in all environments; domain-independent classifiers are introduced by~\citet{wang2020towards} to accomplish target tasks in each known bias situation.

\noindent \textbf{Combating biases with known types.} To alleviate expensive bias annotation costs, some bias-tailored methods relax the demands by requiring only the bias types~\cite{geirhos2018imagenet}. \citet{bahng2020learning} elaborately design specific networks based on the bias types to obtain biased representations on purpose (\textit{e.g.}, using 2D CNNs to extract static bias in action recognition). Then, the debiased representation is learned by encouraging it to be independent of the biased one. \citet{wang2018learning} try to project the model's representation onto the subspace orthogonal to the texture-biased representation. SoftCon~\cite{hong2021unbiased} serves as an extension of BiasCon to handle the cases where only the bias type is available. In addition, the ensemble approach that consists of a bias-type customized biased model and a debiased model is employed in natural language processing as well~\cite{he2019unlearn,clark2019don,cadene2019rubi,UtamaDebias2020,clark2020learning}.

\noindent \textbf{Combating unknown biases.} Despite the effectiveness of the methodologies described above, the assumptions limit their applications, as manually discovering bias types heavily relies on experts' knowledge and labeling bias attributes for each training sample is even more laborious. As a result, recent studies~\cite{le2020adversarial,kim2019multiaccuracy,hashimoto2018fairness} try to obtain debiased models with unknown biases, which are more realistic. \citet{nam2020learning} mine bias-conflicting samples with generalized cross entropy (GCE) loss~\cite{zhang2018generalized} and emphasize them using a designed weight assignment function. \citet{kim2021learning} further synthesize diverse bias-conflicting samples via feature-level data augmentation, whereas \citet{kim2021biaswap} directly generate them with SwapAE~\cite{park2020swapping}. RNF~\cite{du2021fairness} uses the neutralized representations from samples with the same target label but different bias labels (generated by GCE-based biased models, the version that accesses real bias labels is called RNF-GT) to train the classification head alone. Besides GCE loss, feature clustering~\cite{sohoni2020no}, early-stopping~\cite{liu2021just}, forgettable examples~\cite{yaghoobzadeh2021increasing} and limited network capacity~\cite{sanh2020learning,Utama2020TowardsDN} are involved to identify bias-conflicting samples. Furthermore,~\citet{creager2021environment} and~\citet{lahoti2020fairness} alternatively infer dataset partitions and enhance domain-invariant feature learning by min-max adversarial training. In addition to the identify-emphasize paradigm,~\citet{pezeshki2020gradient} introduces a novel regularization method for decoupling feature learning dynamics in order to improve model robustness.

\section{Methodology}
\label{sec:method}

\subsection{Effective bias-conflicting scoring}
\label{sec:det}
Due to the explicit bias information is not available, we try to describe how likely input $x$ is a bias-conflicting sample via the \textbf{bias-conflicting (b-c) score}: $s(x,y)$ $\in$ [0,1], $y \in \{1,2,\cdots,C\}$ stands for the target label. A larger $s(x,y)$ indicates that $x$ is harder to be recognized via unintended decision rules. As models are prone to fitting shortcuts, previous studies~\cite{kim2021biaswap,liu2021just} resort model's output probability on target class to define $s(x,y)$ as $1 - p(y | \dot{f}(x) )$, where $p(k|\dot{f}(x)) = \frac{e^{\dot{f}(x)[k]}}{\sum_{k'=1}^C e^{\dot{f}(x)[k']}}$, $\dot{f}$ is an auxiliary biased model and $\dot{f}(x)[k]$ denotes the $k^{th}$ index of logits $\dot{f}(x)$. Despite this, over-parameterized networks tend to ``memorize" all samples, resulting in low scores for the real bias-conflicting samples as well. To avoid it, we propose the following two strategies. The whole scoring framework is summarized in Alg.~\ref{alg:det} (noting that the ``for'' loop is used for better clarification, it can be avoided in practice).

\begin{algorithm}[t]
\begin{footnotesize}
\caption{\small \textbf{E}ffective bias-\textbf{C}onflicting \textbf{S}coring (ECS)}
\label{alg:det}
\KwIn{$\mathcal{D}$=$\{(x^{i},y^{i})\}_{i=1}^N$; initial models $\dot{f}^0$,$\ddot{f}^0$ and b-c scores $\{s^i\gets 0 \}_{i=1}^N$; loss function $\ell$; threshold $\eta$.}
\For{$t=0$ \KwTo $T-1$}{
$\mathcal{B}$=$\{(x^{j}, y^{j})\}_{j=1}^B$ $\gets$ \text{Batch}($\mathcal{D}$) \tcp{batch size $B$}

$\{p( y^{j} | \dot{f}^{t}(x^{j}) )\}$,$\{p( y^{j} | \ddot{f}^{t}(x^{j}) )\}$$\gets$\text{Forward}($\mathcal{B}$,$\dot{f}^{t}$, $\ddot{f}^{t})$ \; 

$\dot{l}^t \gets 0; \quad \ddot{l}^t \gets 0;$ \tcp{initialize loss}

\For{$j=1$ \KwTo $B$}{
\uIf{$p(y^{j} | \dot{f}(x^{j}))$$>\eta$ and $\ p(y^{j} | \ddot{f}(x^{j}))$$>\eta$}{
$\dot{l}^t \mathrel{+}= \ell(\dot{f}^{t}(x^{j}), y^{j}); \ddot{l}^t \mathrel{+}= \ell(\ddot{f}^{t}(x^{j}), y^{j})$ \; 
}
\uElseIf{$p(y^{j} | \dot{f}(x^{j}))$$>\eta$ and $p(y^{j} | \ddot{f}(x^{j}))$$\leq\eta$}{
$\dot{l}^t \mathrel{-}= \ell(\dot{f}^{t}(x^{j}), y^{j})$ \; 
}
\ElseIf{$p(y^{j} | \dot{f}(x^{j}))$$\leq\eta$ and $p(y^{j} | \ddot{f}(x^{j}))$$>\eta$}{
$\ddot{l}^t \mathrel{-}= \ell(\ddot{f}^{t}(x^{j}), y^{j})$ \; 
}
}
$\dot{f}^{t+1} \gets \text{Update} (\dot{f}^{t}, \frac{\dot{l}^t}{B})$  ;\
$\ddot{f}^{t+1} \gets \text{Update} (\ddot{f}^{t}, \frac{\ddot{l}^t}{B})$ \; 
\If{$(t+1) \% T' =0$}{
\For{$i=1$ \KwTo $N$}{
$s^{i} \mathrel{+}= \frac{T'}{T} [1-\frac{p(y^{i} | \dot{f}^{t+1}(x^{i}))+p(y^{i} | \ddot{f}^{t+1}(x^{i}))}{2}]$ 
}
}
}
\KwOut{the estimated b-c scores $\{s^i\}_{i=1}^N$.}
\end{footnotesize}
\end{algorithm}

\noindent \textbf{Training auxiliary biased models with peer-picking.} Deliberately amplifying the auxiliary model's bias seems to be a promising strategy for better scoring~\cite{nam2020learning}, as heavily biased models can assign high b-c scores to bias-conflicting samples. We achieve this by \textbf{confident-picking} --- only picking samples with confident predictions (which are more like bias-aligned samples) to update auxiliary models. Nonetheless, a few bias-conflicting samples can still be overfitted and the memorization will be strengthened with continuous training. Thus, with the assist of \textbf{peer model}, we propose \textbf{peer-picking}, a co-training-like~\cite{han2018co} paradigm, to train auxiliary biased models (more auxiliary models can get slightly better results but raise costs).

Our method maintains two auxiliary biased models $\dot{f}$ and $\ddot{f}$ simultaneously (identical structure). Considering a training set $\mathcal{D}$ = $\{(x^{i},y^{i})\}^N_{i=1}$ with $B$ samples in each batch, with a threshold $\eta$ $\in$ (0,1), each model divides samples into confident and unconfident groups relying on the output probabilities on target classes. Consequently, four clusters are formed as shown in Fig.~\ref{fig:intro}. For the red cluster ($\mathcal{O}_1$), since two models are confident on them, it is reasonable to believe that they are indeed bias-aligned samples, therefore we pick up them to update model via gradient descent as usual (L7,12 of Alg.\ref{alg:det}). While the gray cluster ($\mathcal{O}_2$), on which both two models are unconfident, will be discarded outright as they might be bias-conflicting samples. The remaining purple clusters ($\mathcal{O}_3$ and $\mathcal{O}_4$) indicate that some samples may be bias-conflicting, but they are memorized by one of auxiliary models. Inspired by the work for handling noisy labels~\cite{pmlr-v119-han20c}, we endeavor to force the corresponding model to forget the memorized suspicious samples via gradient ascent (L9,11,12). We average the output results of the two heavily biased models $\dot{f}$ and $\ddot{f}$ to obtain b-c scores (L15).

\noindent \textbf{Collecting results with epoch-ensemble.} During the early stage of training, b-c scores $\{s^i\}$ ($s^i$:=$s(x^{i}, y^{i})$) of real bias-conflicting samples are usually higher than those of bias-aligned ones, while the scores may be indistinguishable at the end of training due to overfitting. Unfortunately, selecting an optimal moment for scoring is strenuous. To avoid tedious hyper-parameter tuning, we collect results every $T'$ iterations (typically every epoch in practice, \textit{i.e.}, $T'=\lfloor \frac{N}{B} \rfloor$) and adopt the ensemble averages of multiple results as the final b-c scores (L15). We find that the ensemble can alleviate the randomness of a specific checkpoint and achieve superior results without using tricks like early-stopping.

\subsection{Gradients alignment}
\label{sec:ga}
\begin{figure}[t]
\centering
\includegraphics[width=0.5\columnwidth]{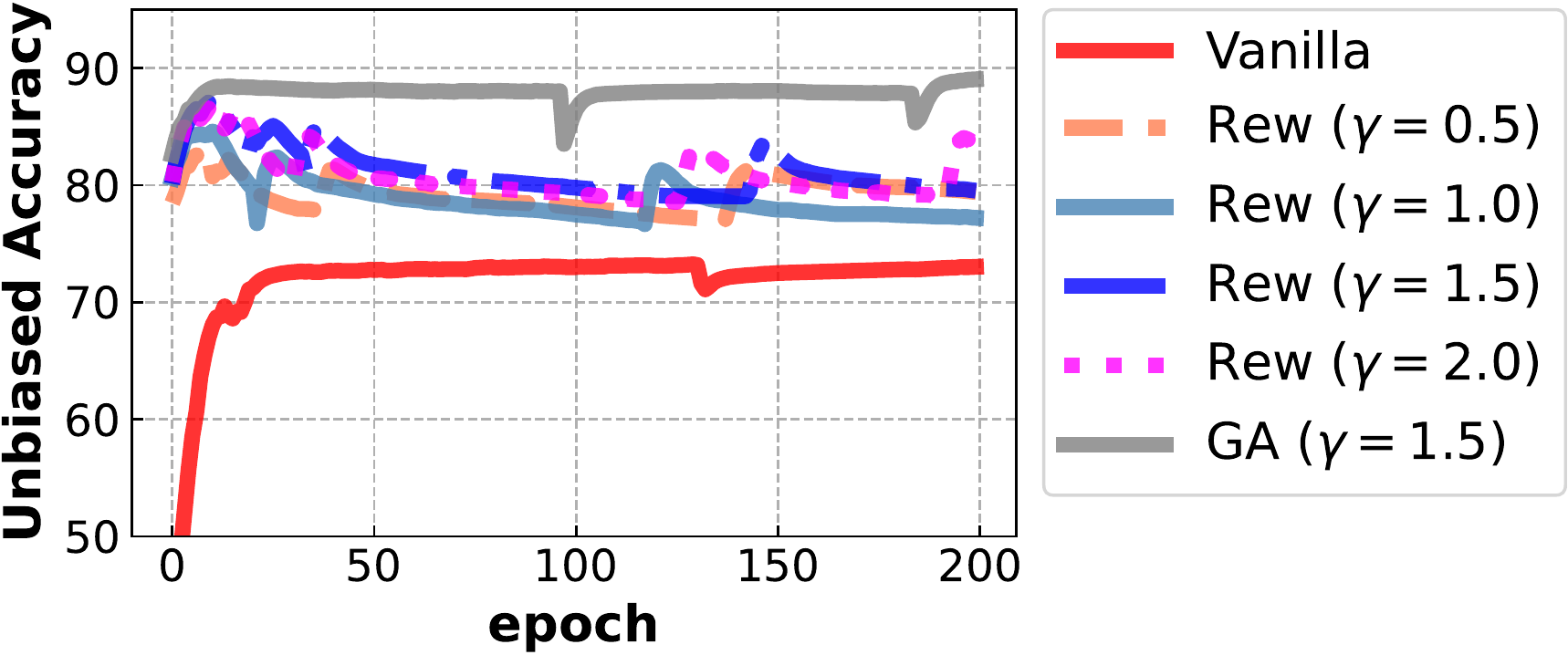} 
\caption{Unbiased accuracy (\%) on Colored MNIST.}
\label{fig:acc_c_mnist}
\end{figure}

Then, we attempt to train the debiased model $f$. We focus on an important precondition of the presence of biased models: the training objective can be achieved through unintended decision rules. To avoid it, one should develop a new learning objective that cannot be accomplished by these rules. The most straightforward inspiration is the use of plain reweighting (Rew) to intentionally rebalance sample contributions from different domains~\cite{sagawa2020investigation}:
\begin{small}
\begin{equation}
\mathcal{R}_{Rew} = \sum_{i=1}^{\underline{N}} \frac{\overline{N}}{\gamma \cdot \underline{N}} \cdot \ell(f(\underline{x}^{i}), y^{i}) + \sum_{j=1}^{\overline{N}} \ell(f(\overline{x}^{j}), y^{j}),
\label{eq:reweight}
\end{equation}
\end{small}where $\overline{N}$ and $\underline{N}$ are the number of bias-conflicting and bias-aligned samples respectively, $\gamma$ $\in$ $(0, \infty)$ is a reserved hyper-parameter to conveniently adjust the tendency: when $\gamma$$\rightarrow$0, models intend to exploit bias-aligned samples more and when $\gamma$$\rightarrow$$\infty$, the behavior is reversed. As depicted in Fig.~\ref{fig:acc_c_mnist}, assisted with reweighting, unbiased accuracy skyrockets in the beginning, indicating that the model tends to learn intrinsic features in the first few epochs, while declines gradually, manifesting that the model is biased progressively (adjusting $\gamma$ can not reverse the tendency).

The above results show that the static ratio between $\overline{N}$ and $\underline{N}$ is not a good indicator to show how balanced the training is, as the influence of samples can fluctuate during training. Accordingly, we are inspired to directly choose gradient statistics as a metric to indicate whether the training is overwhelmed by bias-aligned samples. Let us revisit the commonly used cross-entropy loss:
$
\ell(f(x), y) = -\sum_{k=1}^{C} \mathbb{I}_{k=y} \log p(k|f(x)).
$
For a sample, the gradient on logits $f(x)$ is given by 
\begin{small}
\begin{equation}
\nabla_{f(x)}  \ell(f(x), y) = [
\frac{\partial \ell(f(x), y)}{\partial f(x)[1]},
\frac{\partial \ell(f(x), y)}{\partial f(x)[2]},
\cdots,
\frac{\partial \ell(f(x), y)}{\partial f(x)[C]}
]^{\mathsf{T}}.
\end{equation}  
\end{small}We define the current gradient contribution of sample $x$ as $g(x,y|f)$ = $
|| \nabla_{f(x)}  \ell(f(x), y) ||_1$, which can be efficiently calculated by $\sum_{k=1}^C |\frac{\partial \ell(f(x), y)}{\partial f(x)[k]} |$=$2|\frac{\partial \ell(f(x), y)}{\partial f(x)[y]}|$=$2 - 2p(y|f(x))$. Assuming within the $t^{\text{th}}$ iteration ($t$ $\in$ [0, $T$-1]), the batch is composed of $\underline{B}^t$ bias-aligned and $\overline{B}^t$ bias-conflicting samples ($B$ in total, $\underline{B}^t$ $\gg$ $\overline{B}^t$ under our concerned circumstance). The accumulated gradient contributions generated by bias-aligned samples are denoted as $\underline{g}^t$ = $\sum_{i=1}^{\underline{B}^t} g(\underline{x}^i,y^i|f^t)$, similarly for the contributions of bias-conflicting samples: $\overline{g}^t$.

\begin{figure}[t]
\centering
\includegraphics[width=0.8\columnwidth]{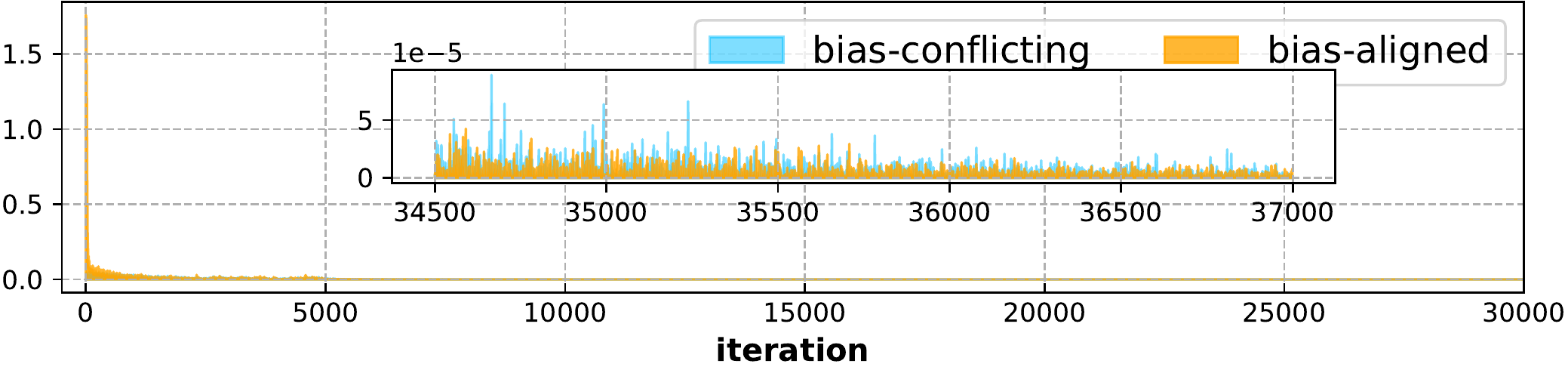}\\
\includegraphics[width=0.8\columnwidth]{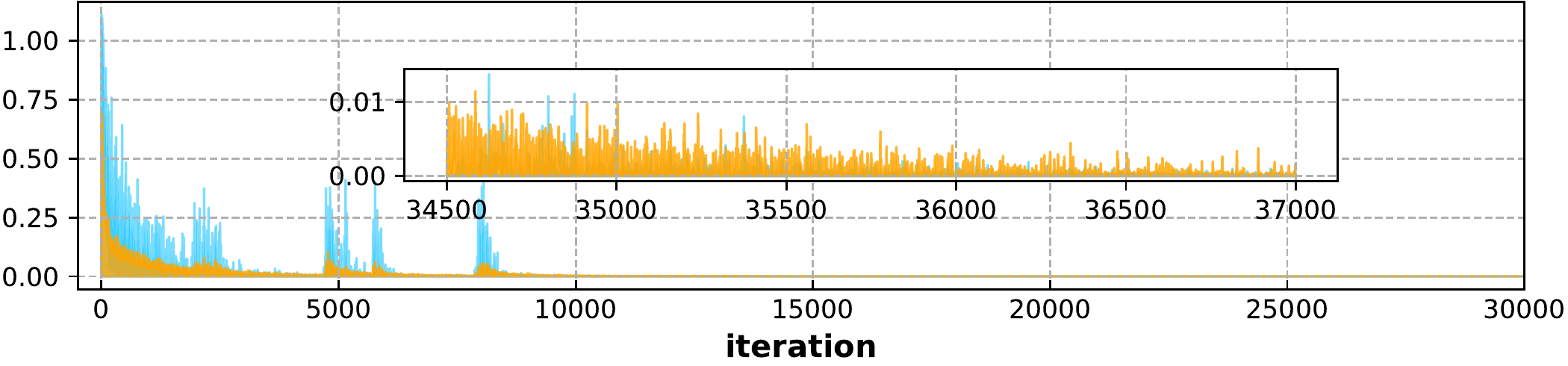}\\
\includegraphics[width=0.8\columnwidth]{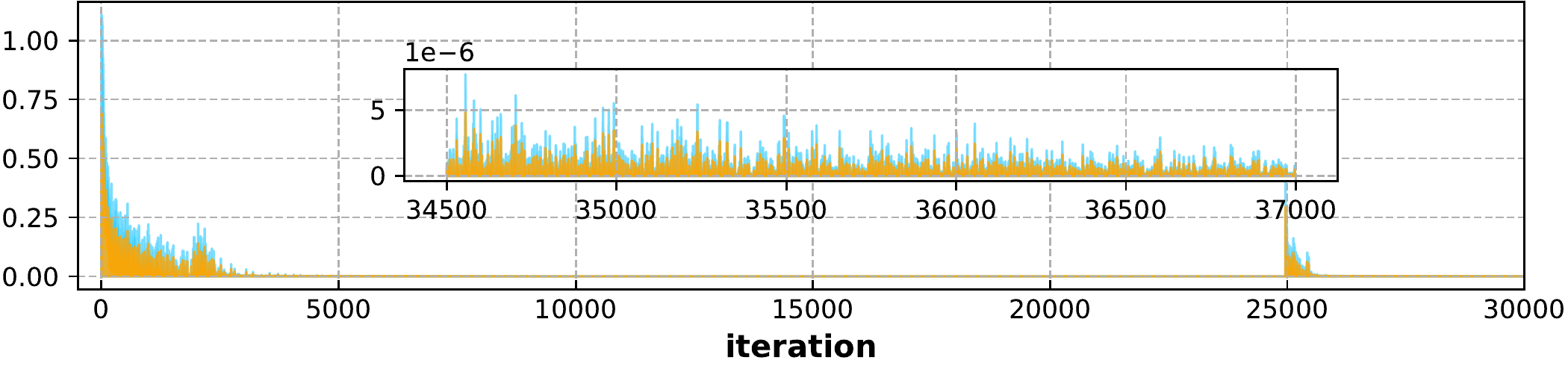}
\caption{Statistics of $\{ \overline{g}^t \}_{t=0}^{T-1}$ and $\{ \underline{g}^t \}_{t=0}^{T-1}$. Vanilla (top), Rew (middle), GA (bottom). Results in the late stage are enlarged and shown in each figure.}
\label{fig:grad_c_mnist}
\end{figure}

We present the statistics of $\{ \overline{g}^t \}_{t=0}^{T-1}$ and $\{ \underline{g}^t \}_{t=0}^{T-1}$ when learning with standard ERM (Vanilla), Eq.~\eqref{eq:reweight} (Rew) in Fig.~\ref{fig:grad_c_mnist}.
For vanilla training, we find the gradient contributions of bias-aligned samples overwhelm that of bias-conflicting samples at the beginning, thus the model becomes biased towards spurious correlations rapidly. Even though at the late stage, the gap in gradient contributions shrinks, it is hard to rectify the already biased model.
For Rew, we find the contributions of bias-conflicting and bias-aligned samples are relatively close at the beginning (compared to those under Vanilla), thus both of them can be well learned. Nonetheless, the bias-conflicting samples are memorized soon due to their small quantity, and the gradient contributions from the bias-conflicting samples become smaller than that of the bias-aligned samples gradually, leading to biased models step by step.

The above phenomena are well consistent with the accuracy curves in Fig.~\ref{fig:acc_c_mnist}, indicating that the gradient statistics can be a useful ``barometer'' to reflect the optimization process. Therefore, the core idea of gradient alignment is to rebalance bias-aligned and bias-conflicting samples according to their currently produced gradient contributions. Within the $t^{\text{th}}$ iteration, We define the contribution ratio $r^t$ as:
\begin{small}
\begin{equation}
r^t = \frac{ \overline{g}^t}
{\gamma \cdot \underline{g}^t} = 
\frac{\sum_{j=1}^{ \overline{B}^t } [1 - p(y^j| f^{t}(\overline{x}^{j}) ) ]}
{\gamma \cdot \sum_{i=1}^{\underline{B}^t} [1 - p(y^i | f^{t}(\underline{x}^{i}) ) ]},
\label{eq:r_ga}
\end{equation}
\end{small}where $\gamma$ plays a similar role as in Rew. Then, with $r^t$, we rescale the gradient contributions derived from bias-aligned samples to achieve alignment with that from bias-conflicting ones, which can be simply implemented by reweighting the learning objective for the $t^{\text{th}}$ iteration:
\begin{small}
\begin{equation}
\mathcal{R}_{GA}^t = \sum_{i=1}^{\underline{B}^t} r^t \cdot  \ell(f^{t}(\underline{x}^{i}), y^{i}) + \sum_{j=1}^{\overline{B}^t} \ell(f^{t}(\overline{x}^{j}), y^{j}),
\label{eq:GA}
\end{equation}
\end{small}\textit{i.e.}, the modulation weight is adaptively calibrated in each iteration. As shown in Eq.~\eqref{eq:r_ga} and~\eqref{eq:GA}, GA only needs negligible computational extra cost (1$\times$ forward and backward as usual, only increases the cost of computing $r^t$). As shown in Fig.~\ref{fig:grad_c_mnist}, GA can dynamically balance the contributions throughout the whole training process. Correspondingly, it obtains optimal and stable predictions as demonstrated in Fig.~\ref{fig:acc_c_mnist} and multiple other challenging datasets in Sec.~\ref{sec:exp}. Noting that as bias-conflicting samples are exceedingly scarce, it is unrealistic to ensure that every class can be sampled in one batch, thus all classes share the same ratio in our design.

\begin{table*}[t]
\centering
\resizebox{\textwidth}{!}{
\begin{tabular}{>{\Large}l | >{\Large}c >{\Large}c >{\Large}c >{\Large}c | >{\Large}c >{\Large}c >{\Large}c >{\Large}c | >{\Large}c >{\Large}c >{\Large}c >{\Large}c | >{\Large}c | >{\Large}c}
\toprule
& \multicolumn{4}{>{\Large}c|}{Colored MNIST} & \multicolumn{4}{>{\Large}c|}{Corrupted CIFAR10$^1$} & \multicolumn{4}{>{\Large}c|}{Corrupted CIFAR10$^2$} & B-Birds  & B-CelebA \\
\multicolumn{1}{>{\Large}c|}{$\rho$} & 95\% & 98\% & 99\% & 99.5\% & 95\% & 98\% & 99\% & 99.5\% & 95\% & 98\% & 99\% & 99.5\% & 95\%  & 99\% \\
\midrule
Vanilla & 85.7$_{\pm 0.1}$  &    73.6$_{\pm 0.5}$   &   60.7$_{\pm 0.6}$   &    45.4$_{\pm 0.8}$   &   44.9$_{\pm 1.0}$   &  30.4$_{\pm 1.0}$   & 22.4$_{\pm 0.8}$  &  17.9$_{\pm 0.9}$   &  42.7$_{\pm 0.9}$   &   27.2$_{\pm 0.6}$  &   20.6$_{\pm 0.5}$ &  17.4$_{\pm 0.8}$ &  77.1$_{\pm 1.5}$ & 77.4$_{\pm 1.6}$\\ 
Focal &  86.7$_{\pm 0.2}$   &   75.8$_{\pm 0.6}$    & 62.4$_{\pm 0.3}$   &   45.9$_{\pm 0.9}$  & 45.5$_{\pm 1.0}$   &    
30.7$_{\pm 1.1}$ &  22.9$_{\pm 1.1}$  &  
17.8$_{\pm 0.5}$  & 41.9$_{\pm 0.5}$  &  26.9$_{\pm 0.5}$  &  21.0$_{\pm 0.6}$   &  17.0$_{\pm 0.2}$   &   78.6$_{\pm 0.7}$  &  78.1$_{\pm 1.0}$  \\ 
  GEORGE  &  87.0$_{\pm 0.5}$   &  76.2$_{\pm 0.9}$  & 62.4$_{\pm 0.6}$    &   46.4$_{\pm 0.2}$  & 44.6$_{\pm 1.0}$  & 29.5$_{\pm 1.0}$  &   21.8$_{\pm 0.3}$   &    17.9$_{\pm 0.6}$   &    44.2$_{\pm 1.9}$  &  27.3$_{\pm 1.6}$   &    20.7$_{\pm 1.2}$   &    17.7$_{\pm 0.3}$  & 79.3$_{\pm 0.9}$   & 78.2$_{\pm 0.9}$  \\ 
 LfF   &  88.2$_{\pm 0.9}$ & 86.7$_{\pm 0.6}$  &   80.3$_{\pm 1.2}$ & 73.2$_{\pm 0.9}$   &   59.6$_{\pm 0.8}$ & 50.4$_{\pm 0.5}$ &  42.9$_{\pm 2.8}$    &  34.6$_{\pm 2.3}$  &   58.5$_{\pm 0.8}$ &   49.0$_{\pm 0.4}$   &   42.2$_{\pm 1.1}$ & 33.4$_{\pm 1.2}$ & 80.4$_{\pm 1.1}$ &  84.4$_{\pm 1.5}$ \\ 
 DFA    &  89.8$_{\pm 0.2}$  &  86.9$_{\pm 0.4}$ &  81.8$_{\pm 1.1}$ & 74.1$_{\pm 0.8}$ & 58.2$_{\pm 1.8}$  & 50.0$_{\pm 2.3}$   & 41.8$_{\pm 4.7}$  &  35.6$_{\pm 4.6}$  & 58.6$_{\pm 0.2}$ & 48.7$_{\pm 0.6}$  & 41.5$_{\pm 2.2}$    & 35.2$_{\pm 1.9}$  &79.5$_{\pm 0.7}$  & 84.3$_{\pm 0.6}$ \\ 
 SD    &  86.7$_{\pm 0.3}$   &  73.9$_{\pm 0.2}$   &  59.7$_{\pm 0.5}$  &   42.4$_{\pm 1.1}$   &   43.1$_{\pm 0.5}$  &   28.6$_{\pm 1.5}$   &   21.6$_{\pm 0.9}$   &   17.7$_{\pm 0.6}$   &   41.4$_{\pm 0.3}$    &   27.0$_{\pm 0.8}$    &    20.0$_{\pm 0.2}$   &  17.5$_{\pm 0.3}$  & 76.8$_{\pm 1.3}$ &  77.8$_{\pm 1.1}$  \\ 
  ECS+Rew  &   91.8$_{\pm 0.2}$  &   88.6$_{\pm 0.7}$  &  84.2$_{\pm 0.3}$  & 78.9$_{\pm 0.9}$  & 58.5$_{\pm 0.0}$  &  47.5$_{\pm 0.6}$  &  38.6$_{\pm 1.1}$  & 33.4$_{\pm 1.2}$  &  61.4$_{\pm 0.7}$   &  53.2$_{\pm 0.3}$  &  47.4$_{\pm 1.2}$  &  40.3$_{\pm 0.6}$  & 82.7$_{\pm 0.7}$   & 88.3$_{\pm 0.4}$  \\ 
 ECS+ERew  & 91.0$_{\pm 0.2}$  &  87.5$_{\pm 0.2}$  &  81.4$_{\pm 0.9}$ &   71.3$_{\pm 2.2}$  &    59.8$_{\pm 0.5}$   &   47.9$_{\pm 1.0}$    &   38.5$_{\pm 0.2}$    &   30.2$_{\pm 1.3}$    &  62.2$_{\pm 0.5}$  &     51.1$_{\pm 0.2}$  &   41.4$_{\pm 0.9}$    &  25.9$_{\pm 1.6}$  &  84.9$_{\pm 0.9}$ & 80.5$_{\pm 0.6}$  \\ 
 ECS+PoE   &   80.2$_{\pm 1.5}$    &  75.4$_{\pm 1.4}$ & 64.4$_{\pm 2.7}$  & 50.0$_{\pm 3.0}$  & 54.4$_{\pm 0.2}$   &     48.7$_{\pm 1.3}$  &   \textbf{45.6}$_{\pm 1.3}$  &  \textbf{42.7}$_{\pm 0.8}$  & 47.9$_{\pm 0.8}$      &     40.3$_{\pm 1.3}$  &    36.8$_{\pm 2.5}$   & \textbf{42.4}$_{\pm 2.3}$ & 85.8$_{\pm 0.6}$ &  81.1$_{\pm 0.1}$ \\ 
  ECS+GA  &  \textbf{92.1}$_{\pm 0.1}$  &   \textbf{89.5}$_{\pm 0.4}$  &  \textbf{86.4}$_{\pm 0.5}$   &  \textbf{79.9}$_{\pm 0.8}$  &  \textbf{61.0}$_{\pm 0.1}$   &  \textbf{51.7}$_{\pm 0.5}$   &  42.6$_{\pm 0.7}$ &   35.0$_{\pm 0.5}$  &  \textbf{64.1}$_{\pm 0.3}$  & \textbf{57.0}$_{\pm 0.6}$    & \textbf{50.0}$_{\pm 1.5}$   & 41.8$_{\pm 0.8}$ &  \textbf{86.1}$_{\pm 0.5}$   &  \textbf{89.5}$_{\pm 0.5}$\\ 
\midrule
\midrule
$^\dag$REBIAS &  85.5$_{\pm 0.6}$  &  74.0$_{\pm 0.7}$  &  61.1$_{\pm 0.8}$  &  44.5$_{\pm 0.4}$ & 44.8$_{\pm 0.3}$   &  29.9$_{\pm 0.7}$ &  22.4$_{\pm 1.1}$   &  17.7$_{\pm 0.3}$  &  41.5$_{\pm 1.0}$ & 27.0$_{\pm 0.6}$  & 20.6$_{\pm 0.6}$   & 17.9$_{\pm 0.3}$ & 77.5$_{\pm 0.6}$  & 78.1$_{\pm 1.2}$  \\ 
$^\dag$Rew  &   91.5$_{\pm 0.0}$  &   87.9$_{\pm 0.4}$     &  83.8$_{\pm 0.6}$  & 77.6$_{\pm 0.7}$  & 59.1$_{\pm 0.2}$  &  48.9$_{\pm 0.8}$  &  40.4$_{\pm 0.4}$  & 33.4$_{\pm 1.4}$  &  61.1$_{\pm 0.2}$   &  53.1$_{\pm 0.8}$  &  46.9$_{\pm 1.1}$  &  41.2$_{\pm 0.6}$  & 86.0$_{\pm 0.4}$   & 90.7$_{\pm 0.4}$  \\ 
$^\dag$RNF-GT &   84.3$_{\pm 4.1}$  &   75.9$_{\pm 3.6}$     &  66.3$_{\pm 8.2}$  & 59.1$_{\pm 5.7}$  & 52.1$_{\pm 0.7}$  &  39.1$_{\pm 1.2}$  &  30.6$_{\pm 1.3}$  & 22.2$_{\pm 0.4}$  &  50.3$_{\pm 1.0}$   &  34.9$_{\pm 0.5}$  &  27.9$_{\pm 0.6}$  &  19.8$_{\pm 0.4}$  & 81.2$_{\pm 1.3}$   & 85.1$_{\pm 2.7}$  \\ 
$^\dag$BiasCon  &   90.9$_{\pm 0.1}$  &   86.7$_{\pm 0.1}$     &  83.0$_{\pm 0.0}$  & 79.0$_{\pm 1.5}$  & 59.0$_{\pm 0.6}$  &  48.6$_{\pm 0.6}$  &  39.0$_{\pm 0.4}$  & 32.4$_{\pm 0.3}$  &  60.0$_{\pm 0.3}$   &  49.9$_{\pm 0.3}$  &  43.0$_{\pm 0.4}$  &  37.4$_{\pm 0.8}$  & 84.1$_{\pm 0.6}$   & 90.4$_{\pm 1.2}$  \\ 
$^\dag$GA  &  92.4$_{\pm 0.3}$ &  89.1$_{\pm 0.2}$ &  85.7$_{\pm 0.4}$ &  80.4$_{\pm 0.5}$    &  61.5$_{\pm 0.8}$   &  52.9$_{\pm 0.3}$ & 43.5$_{\pm 1.6}$  & 33.9$_{\pm 0.8}$ & 64.5$_{\pm 0.2}$  & 56.9$_{\pm 0.2}$  & 51.1$_{\pm 0.3}$ & 43.6$_{\pm 0.8}$ & 87.9$_{\pm 0.5}$  &  92.3$_{\pm 0.2}$ \\ 
\bottomrule
\end{tabular}%
}
\caption{Overall unbiased accuracy (\%) and standard deviation over three runs. Best results with unknown biases are shown in bold. $^\dag$ indicates that the method requires prior knowledge regarding bias.}
\label{tab:overall_acc}%
\end{table*}

To handle unknown biases, we simply utilize the estimated b-c score $\{s^i\}_{i=1}^N$ and a threshold $\tau$ to assign input $x$ as bias-conflicting ($s(x,y)$ $\geq$ $\tau$) or bias-aligned ($s(x,y)$ \textless $\tau$) here. For clarity, GA with the pseudo annotations (bias-conflicting or bias-aligned) produced by ECS will be denoted as `ECS+GA' (similarly, `ECS+$\triangle$' represents combining ECS with method $\triangle$).

\section{Experiments}
\label{sec:exp}

\subsection{Experimental Settings}
\label{sec:settings}

\textbf{Datasets.} We mainly conduct experiments on five benchmark datasets. For Colored MNIST (C-MNIST), the task is to recognize digits (0-9), in which the images of each target class are dyed by the corresponding color with probability $\rho$ $\in$ $\{95\%,98\%,99\%,99.5\%\}$ and by other colors with probability $1-\rho$ (a higher $\rho$ indicates more severe biases).
Similarly, for Corrupted CIFAR10, each object class in it holds a spurious correlation with a corruption type. Two sets of corruption protocols are utilized, leading to two biased datasets~\cite{nam2020learning}: Corrupted CIFAR10$^1$ and CIFAR10$^2$ (C-CIFAR10$^1$, C-CIFAR10$^2$) with $\rho \in \{95\%,98\%,99\%,99.5\%\}$.
In Biased Waterbirds (B-Birds), ``waterbirds'' and ``landbirds'' are highly correlated with ``wet'' and ``dry'' habitats (95\% bias-aligned samples, \textit{i.e.}, $\rho$=95\%). Consequently, the task aiming to distinguish images as ``waterbird" or ``landbird" can be influenced by background. To focus on the debiasing problem, we balance the number of images per class.
In Biased CelebA (B-CelebA), blond hair is predominantly found on women, whereas non-blond hair mostly appears on men ($\rho$=99\%). When the goal is to classify the hair color as ``blond" or ``non-blond", the information of gender (``male'' or ``female'' in this dataset) can be served as a shortcut~\cite{nam2020learning}.

\noindent \textbf{Compared methods.} We choose various methods for comparison: standard ERM (Vanilla), Focal loss~\cite{lin2017focal}, plain reweighting~\cite{sagawa2020investigation} (Rew, ECS+Rew), REBIAS~\cite{bahng2020learning}, BiasCon~\cite{hong2021unbiased}, RNF-GT~\cite{du2021fairness}, GEORGE~\cite{sohoni2020no}, LfF~\cite{nam2020learning}, DFA~\cite{kim2021learning}, SD~\cite{pezeshki2020gradient}, ERew~\cite{clark2019don} and PoE~\cite{clark2019don} (ECS+ERew, ECS+PoE)\footnote{As the auxiliary biased models used in ERew and PoE are designed for NLP tasks, here, we combine our ECS with them.}. Among them, REBIAS requires bias types; Rew, BiasCon, RNF-GT, and GA are performed with real conflicting/aligned annotations\footnote{As stated in the original papers, BiasCon and RNF-GT have variations that do not require real annotations assist with various auxiliary biased models. We only provide the upper bound of these strategies when combating unknown biases, as we found that the auxiliary models have a significant impact on the outcomes.}.

\noindent \textbf{Evaluation metrics.} Following~\citet{nam2020learning}, we mainly report the overall unbiased accuracy, alongside the fairness performance in terms of DP and EqOdd~\cite{reddy2021benchmarking}.

\begin{table*}
\centering
\resizebox{\textwidth}{!}{
\begin{tabular}{ >{\Large}l | >{\Large}c >{\Large}c >{\Large}c | >{\Large}c >{\Large}c >{\Large}c | >{\Large}c >{\Large}c >{\Large}c | >{\Large}c >{\Large}c >{\Large}c | >{\Large}c >{\Large}c >{\Large}c }
\toprule
 &  \multicolumn{3}{c|}{Colored MNIST ($\rho$=98\%)} & \multicolumn{3}{c|}{Corrupted CIFAR-10$^1$ ($\rho$=98\%)} & \multicolumn{3}{c|}{Corrupted CIFAR-10$^2$ ($\rho$=98\%)} & \multicolumn{3}{c|}{Biased Waterbirds} & \multicolumn{3}{c}{Biased CelebA} \\
& best $\uparrow$ &  last $\uparrow$ & $\Delta$ $\downarrow$
&  best $\uparrow$ &  last $\uparrow$ & $\Delta$ $\downarrow$
&  best $\uparrow$ &  last $\uparrow$  & $\Delta$ $\downarrow$
&  best $\uparrow$ &  last $\uparrow$  & $\Delta$ $\downarrow$
&  best $\uparrow$ &  last  $\uparrow$ & $\Delta$ $\downarrow$ \\
\midrule
LfF & 86.7$_{\pm 0.6}$ & 75.1$_{\pm 0.3}$ & 11.6$_{\pm 0.8}$ & 50.4$_{\pm 0.5}$ & 49.4$_{\pm 0.7}$ & 1.0$_{\pm 0.5}$ & 49.0$_{\pm 0.4}$ & 47.3$_{\pm 0.2}$ & 1.7$_{\pm 0.4}$ & 80.4$_{\pm 1.1}$ & 76.9$_{\pm 2.0}$ & 3.5$_{\pm 3.1}$ & 84.4$_{\pm 1.5}$ & 61.0$_{\pm 1.2}$ & 23.4$_{\pm 2.6}$ \\ 
DFA & 86.9$_{\pm 0.4}$ & 81.2$_{\pm 1.9}$ & 5.8$_{\pm 1.6}$ & 50.0$_{\pm 2.3}$ & 47.9$_{\pm 1.8}$ & 2.1$_{\pm 1.2}$ & 48.7$_{\pm 0.6}$ & 46.0$_{\pm 1.2}$ & 2.7$_{\pm 0.8}$ & 79.5$_{\pm 0.7}$ & 74.5$_{\pm 1.1}$ & 5.0$_{\pm 1.4}$ & 84.3$_{\pm 0.6}$ & 73.2$_{\pm 3.7}$ & 11.0$_{\pm 4.2}$ \\ 
ECS+Rew & 88.6$_{\pm 0.7}$ & 79.6$_{\pm 0.2}$ & 9.0$_{\pm 0.5}$ & 47.5$_{\pm 0.6}$ & 42.6$_{\pm 1.1}$ & 5.0$_{\pm 0.9}$ & 53.2$_{\pm 0.3}$ & 48.8$_{\pm 1.1}$ & 4.4$_{\pm 0.8}$ & 82.7$_{\pm 0.7}$ & 77.2$_{\pm 0.6}$ & 5.5$_{\pm 0.3}$ & 88.3$_{\pm 0.4}$ & 80.5$_{\pm 5.9}$ & 7.8$_{\pm 5.8}$ \\ 
ECS+GA & \textbf{89.5}$_{\pm 0.4}$ & \textbf{88.9}$_{\pm 1.1}$ & \textbf{0.6}$_{\pm 0.7}$ & \textbf{51.7}$_{\pm 0.5}$ & \textbf{50.8}$_{\pm 1.0}$ & \textbf{0.9}$_{\pm 0.5}$ & \textbf{57.0}$_{\pm 0.6}$ & \textbf{55.4}$_{\pm 2.6}$ & \textbf{1.6}$_{\pm 2.1}$ & \textbf{86.1}$_{\pm 0.5}$ & \textbf{85.5}$_{\pm 0.9}$ & \textbf{0.6}$_{\pm 0.6}$ & \textbf{89.5}$_{\pm 0.5}$ & \textbf{87.4}$_{\pm 1.8}$ & \textbf{2.1}$_{\pm 1.4}$ \\ 
\midrule
\midrule
$^\dag$Rew & 87.9$_{\pm 0.4}$ & 79.2$_{\pm 0.2}$ & 8.7$_{\pm 0.3}$ & 48.9$_{\pm 0.8}$ & 45.3$_{\pm 1.8}$ & 3.6$_{\pm 1.4}$ & 53.1$_{\pm 0.8}$ & 47.5$_{\pm 1.8}$ & 5.6$_{\pm 2.1}$ & 86.0$_{\pm 0.4}$ & 78.8$_{\pm 0.5}$ & 7.3$_{\pm 0.2}$ & 90.7$_{\pm 0.4}$ & 82.8$_{\pm 5.1}$ & 7.8$_{\pm 4.9}$ \\ 
$^\dag$BiasCon & 86.7$_{\pm 0.1}$ & 79.4$_{\pm 0.7}$ & 7.3$_{\pm 0.8}$ & 48.6$_{\pm 0.6}$ & 40.9$_{\pm 0.3}$ & 7.7$_{\pm 0.9}$ & 49.9$_{\pm 0.3}$ & 42.0$_{\pm 1.1}$ & 7.9$_{\pm 0.9}$ & 84.1$_{\pm 0.6}$ & 78.5$_{\pm 1.1}$ & 5.7$_{\pm 0.6}$ & 90.4$_{\pm 1.2}$ & 75.2$_{\pm 1.2}$ & 15.2$_{\pm 2.3}$ \\ 
$^\dag$GA & 89.1$_{\pm 0.2}$ & 88.6$_{\pm 0.3}$ & 0.4$_{\pm 0.4}$ & 52.9$_{\pm 0.3}$ & 49.9$_{\pm 1.6}$ & 2.9$_{\pm 1.8}$ & 56.9$_{\pm 0.2}$ & 55.8$_{\pm 0.3}$ & 1.1$_{\pm 0.3}$ & 87.9$_{\pm 0.5}$ & 87.7$_{\pm 0.5}$ & 0.2$_{\pm 0.1}$ & 92.3$_{\pm 0.2}$ & 91.8$_{\pm 0.4}$ & 0.4$_{\pm 0.2}$ \\ 
\bottomrule
\end{tabular}%
}
\caption{Overall accuracy of the best epoch, the last epoch, and their difference (\%).}
\label{tab:last}%
\end{table*}

\noindent \textbf{Implementation.} The studies for the previous debiasing approaches are usually conducted with varying network architectures and training schedules. We run the representative methods with identical configurations to make fair comparisons. We use an MLP with three hidden layers (each hidden layer comprises 100 hidden units) for C-MNIST, except for the biased models in REBIAS (using CNN). ResNet-20~\cite{he2016deep} is employed for C-CIFAR10$^1$ and C-CIFAR10$^2$. ResNet-18 is utilized for B-Birds and B-CelebA.

\subsection{Main results}
\label{sec:quan_com}

\noindent \textbf{The proposed method achieves better performance than others.} The overall unbiased accuracy is reported in Tab.~\ref{tab:overall_acc}. Vanilla models commonly fail to produce acceptable results on unbiased test sets, and the phenomenon is aggravated as $\rho$ goes larger. Different debiasing methods moderate bias propagation with varying degrees of capability. When compared to other SOTA methods, the proposed approach achieves competitive results on C-CIFAR10$^1$ and noticeable improvements on other datasets across most values of $\rho$. For instance, the vanilla model trained on C-CIFAR10$^2$ ($\rho$=99\%) only achieves 20.6\% unbiased accuracy, indicating that the model is heavily biased. While, ECS+GA leads to 50.0\% accuracy, and exceeds other prevailing debiasing methods by 3\%-30\%. When applied to the real-world dataset B-CelebA, the proposed scheme also shows superior results, demonstrating that it can effectively deal with subtle actual biases. Though the main purpose of this work is to combat unknown biases, we find GA also achieves better performance compared to the corresponding competitors when the prior information is available.

\noindent \textbf{Plain reweighting is an important baseline.} We find Rew (and ECS+Rew) can achieve surprising results compared with recent SOTA methods, while it is overlooked by some studies. The results also indicate that explicitly balancing the contributions is extremely important and effective.

\noindent \textbf{Early-stopping is not necessary for GA to select models.} Plain reweighting requires strong regularizations such as early-stopping to produce satisfactory results~\cite{byrd2019effect,Sagawa*2020Distributionally}, implying that the results are not stable. Due to the nature of combating unknown biases, the unbiased validation set is not available, thus recent studies choose to report the best results among epochs~\cite{nam2020learning,kim2021learning} for convenient comparison. We follow this evaluation protocol in Tab.~\ref{tab:overall_acc}. However, in the absence of prior knowledge, deciding when to stop can be troublesome, thus some results in Tab.~\ref{tab:overall_acc} are excessively optimistic. We claim that if the network can attain dynamic balance throughout the training phase, such early-stopping may not be necessary. We further provide the last epoch results in Tab.~\ref{tab:last} to validate it. We find that some methods suffer from serious performance degradation. On the contrary, GA achieves steady results (with the same and fair training configurations). In other words, our method shows superiority under two model selection strategies simultaneously.

\noindent \textbf{The proposed method has strong performance on fairness metrics as well.} As shown in Tab.~\ref{tab:fairness}, the proposed method also obtains significant improvement in terms of DP and EqOdd. These results further demonstrate that the proposed method is capable of balancing bias-aligned and bias-conflicting samples, as well as producing superior and impartial results.

\subsection{Analysis and discussion}
\label{sec:further_ana}

\noindent \textbf{ECS shows superior ability to mine bias-conflicting samples.} We separately verify the effectiveness of each component of ECS on C-MNIST ($\rho$=98\%) and B-CelebA. A good bias-conflicting scoring method should prompt superior precision-recall curves for the mined bias-conflicting samples, \textit{i.e.}, give real bias-conflicting (aligned) samples high (low) scores. Therefore, we provide the average precision (AP) in Tab.~\ref{tab:variants}. When comparing \#0, \#4, \#5, and \#6, we observe that epoch-ensemble, confident-picking and peer model all can improve the scoring method. In addition, as shown in Tab.~\ref{tab:overall_acc}, ECS+GA achieves results similar to GA with the help of ECS; ERew, PoE, and Rew combined with ECS also successfully alleviate biases to some extent, demonstrating that the proposed ECS is feasible, robust, and can be adopted in a variety of debiasing approaches.

\begin{figure}[htbp]
\centering
\begin{minipage}{.7\columnwidth}
   \includegraphics[width=0.49\columnwidth]{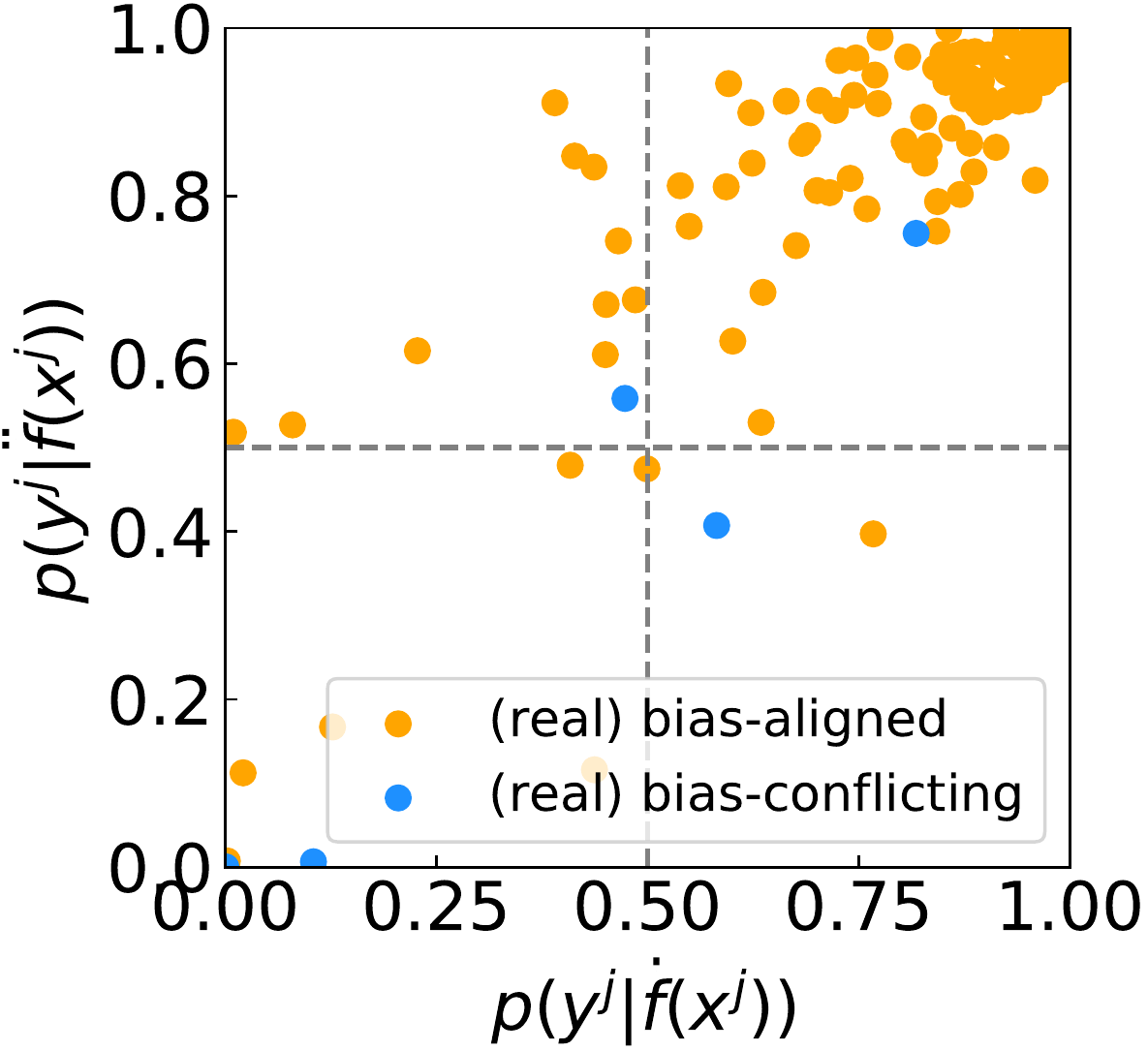}
\includegraphics[width=0.49\columnwidth]{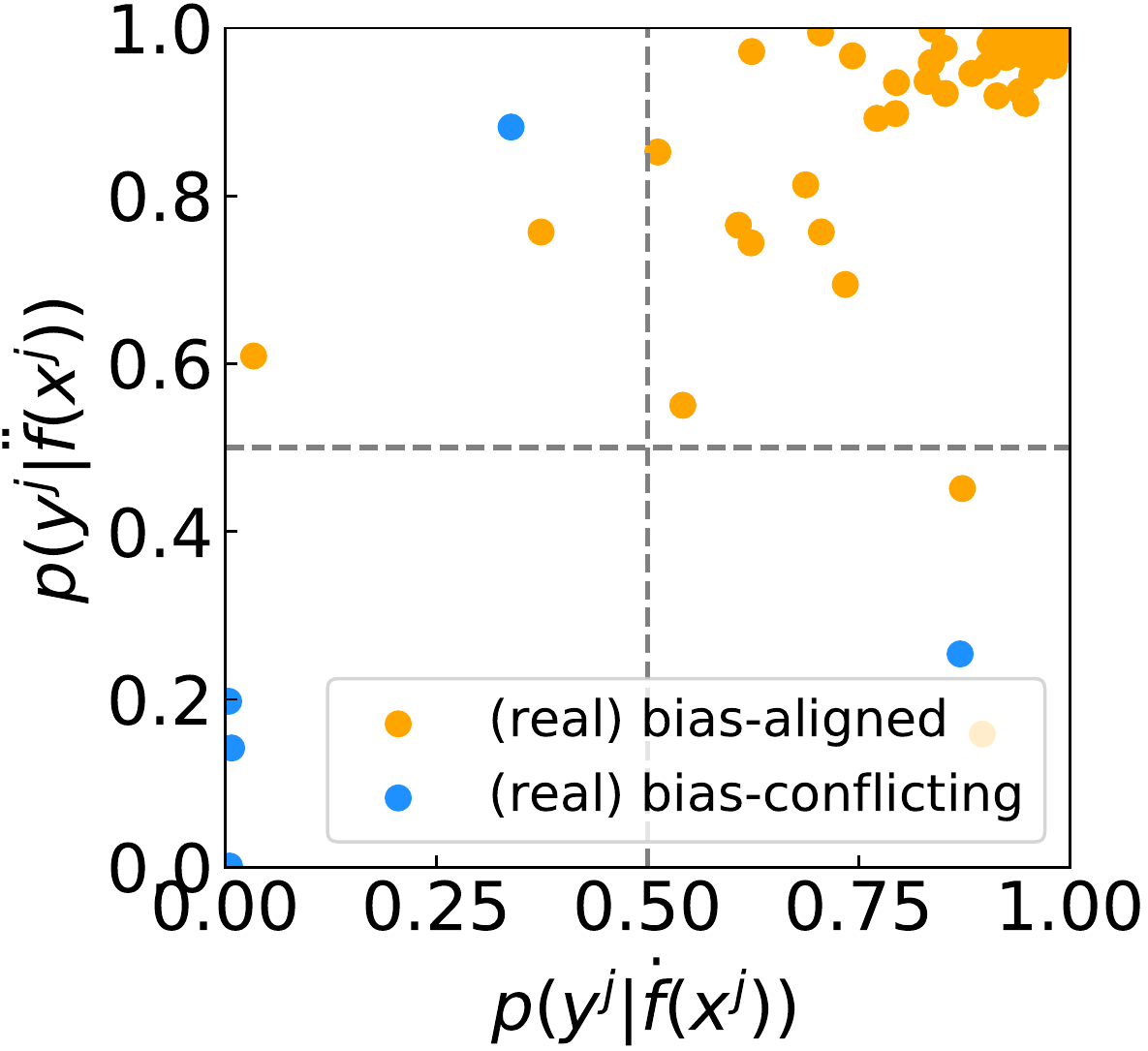} 
\end{minipage}
\begin{minipage}{.28\columnwidth}
\caption{Visualizations of the predictions of peer models (during training).}
\label{fig:peer_vis}
\end{minipage}
\end{figure}

We further compare the methods: \#1 collecting results with early-stopping in JTT~\cite{liu2021just}, \#2 training auxiliary biased model with GCE loss in LfF (and \#3 collecting results with epoch-ensemble on top of it). When comparing \#1 and \#4, both early-stopping and epoch-ensemble can reduce the overfitting to bias-conflicting samples when training biased models, yielding more accurate scoring results. However, early-stopping is laborious to tune~\cite{liu2021just}, whereas epoch-ensemble is more straightforward and robust. From \#2 and \#3, we see that epoch-ensemble can also enhance other strategies. Comparing \#3 and \#5, GCE loss is helpful, while confident-picking gains better results. Noting that though co-training with peer model raises some costs, it is not computationally complex and can yield significant benefits (\#6), and even without peer model, \#5 still outperform previous ways. Peer models are expected to better prevent bias-conflicting samples from affecting the training, so we can get better auxiliary biased models. Though the only difference between peer models is initialization in our experiments, as DNNs are highly nonconvex, different initializations can lead to different local optimal~\cite{han2018co}. We provide the visualizations of the predictions of peer models (during training) in Fig.~\ref{fig:peer_vis}.

\begin{table}[t]
\centering
\begin{minipage}[c]{.48\columnwidth}
\resizebox{\columnwidth}{!}{
\begin{tabular}{lcc}
\toprule
      & DP $\uparrow$   & EqOdd $\uparrow$ \\
\midrule
Vanilla & 0.57$_{\pm 0.01}$ & 0.57$_{\pm 0.01}$ \\
LfF   & 0.63$_{\pm 0.03}$ & 0.61$_{\pm 0.03}$ \\
DFA   & 0.55$_{\pm 0.01}$ & 0.55$_{\pm 0.02}$ \\
ECS+Rew & 0.61$_{\pm 0.02}$ & 0.60$_{\pm 0.01}$ \\
ECS+GA & \textbf{0.99}$_{\pm 0.01}$ & \textbf{0.99}$_{\pm 0.01}$ \\
\bottomrule
\end{tabular}%
}
\end{minipage}
\hspace{1mm}
\begin{minipage}[c]{.48\columnwidth}
\resizebox{\columnwidth}{!}{
\begin{tabular}{lcc}
\toprule
      & DP $\uparrow$  & EqOdd $\uparrow$ \\
\midrule
Vanilla & 0.43$_{\pm 0.01}$ & 0.43$_{\pm 0.02}$ \\
LfF   & \textbf{0.80}$_{\pm 0.06}$ & 0.76$_{\pm 0.07}$ \\
DFA   & 0.69$_{\pm 0.01}$ & 0.76$_{\pm 0.06}$ \\
ECS+Rew & 0.59$_{\pm 0.01}$ & 0.68$_{\pm 0.18}$ \\
ECS+GA & 0.73$_{\pm 0.02}$ & \textbf{0.91}$_{\pm 0.02}$ \\
\bottomrule
\end{tabular}%
}
\end{minipage}
\caption{Performance in terms of DP and EqOdd on Biased Waterbirds (left) and Biased CelebA (right).}
\label{tab:fairness}%
\end{table}

\begin{table}
\centering 
\resizebox{\columnwidth}{!}{
\begin{tabular}{l|>{\Large}l|>{\Large}c >{\Large}c >{\Large}c|>{\Large}c >{\Large}c}
\toprule
&  & Epoch-Ensemble & Confident-Picking & Peer Model  & C-MNIST & B-CelebA \\
\midrule
\#0   & VM &  &   & & 27.0  & 13.3  \\
\#1   & ES (in JTT) &  &   &   & 45.6  & 47.9  \\
\#2   & GCE (in LfF) &   &  &   & 37.0  & 27.8  \\
\#3   & GCE + EE & \checkmark &  &   & 89.3  & 52.1  \\
\midrule
\#4   & ECS (Ours) & \checkmark &   &   & 53.8  & 46.5  \\
\#5   & ECS (Ours)  & \checkmark & \checkmark &  & 95.0  & 61.5  \\
\#6   & ECS (Ours)   & \checkmark & \checkmark & \checkmark & \textbf{98.8}  & \textbf{67.6}  \\
\bottomrule
\end{tabular}%
}
\captionof{table}{Average Precision (\%) of the mined bias-conflicting samples. VM represents scoring with vanilla model.}
\label{tab:variants}%
\end{table}

\noindent \textbf{Why does GA outperform counterparts?} Focal loss, LfF, DFA, and ERew just reweight a sample with the information from itself (individual information), different from them, GA, as well as Rew, use global information within one batch to obtain modulation weight. Correspondingly, the methods based on individual sample information can not maintain the contribution balance between bias-aligned and bias-conflicting samples, which is crucial for this problem as presented in Sec.~\ref{sec:quan_com}.
Different from the static rebalance method Rew, we propose a dynamic rebalance training strategy with aligned gradient contributions throughout the learning process, which enforces models to dive into intrinsic features instead of spurious correlations. Learning with GA, as demonstrated in Fig.~\ref{fig:acc_c_mnist} and Tab.~\ref{tab:last}, produces improved results with no degradation. The impact of GA on the learning trajectory presented in Fig.~\ref{fig:grad_c_mnist} also shows that GA can schedule the optimization processes and take full advantage of the potential of different samples.
Moreover, unlike the methods for class imbalance~\cite{cui2019class,tan2021equalization,zhao2020maintaining}, we try to rebalance the contributions of implicit groups rather than explicit categories.

\noindent \textbf{The method can manage multiple biases.} Most debiasing studies~\cite{nam2020learning,kim2021learning} only discussed single bias. However, there may be multiple biases, which are more difficult to analyze. To study the multiple biases, we adopt the Multi-Color MNIST dataset~\cite{Li_2022_ECCV} which holds two bias attributes: left color ($\rho$=99\%) and right color ($\rho$=95\%). In such training sets, though it seems more intricate to group a sample as bias-aligned or bias-conflicting (as a sample can be aligned or conflicting w.r.t. left color bias or right color bias separately), we still simply train debiased models with GA based on the b-c scores obtained via ECS. We evaluate ECS+GA on four test groups separately and present them in Tab.~\ref{tab:multi_bias}. We find the proposed method also can manage the multi-bias situation.

\begin{table}[t]
\centering
\resizebox{\columnwidth}{!}{
\begin{tabular}{>{\Large}l|>{\Large}c >{\Large}c >{\Large}c >{\Large}c|>{\Large}c}
\toprule
w.r.t. left color bias & aligned & aligned & conflicting & conflicting & \multirow{2}[2]{*}{Avg.} \\
w.r.t. right color bias & aligned & conflicting & aligned & conflicting &  \\
\midrule
LfF$^{\S}$~\cite{nam2020learning}   & 99.6  & 4.7   & \textbf{98.6}  & 5.1   & 52.0  \\
PGI$^{\S}$~\cite{ahmed2020systematic}   & 98.6  & 82.6  & 26.6  & 9.5   & 54.3  \\
EIIL$^{\S}$~\cite{creager2021environment}  & \textbf{100.0}  & \textbf{97.2}  & 70.8  & 10.9  & 69.7  \\
DebiAN$^{\S}$~\cite{Li_2022_ECCV} & \textbf{100.0}  & 95.6  & 76.5  & 16.0  & 72.0  \\
\midrule
ECS+GA  & \textbf{100.0}  & 89.7  & 96.1  & \textbf{24.3}  & \textbf{77.5}  \\
\bottomrule
\end{tabular}%
}
\caption{Accuracies (\%) on four test groups of Multi-Color MNIST. $^{\S}$states the reported results from DebiAN.}
\label{tab:multi_bias}%
\end{table}%

\begin{figure}[t]
\centering
\includegraphics[width=0.32\columnwidth]{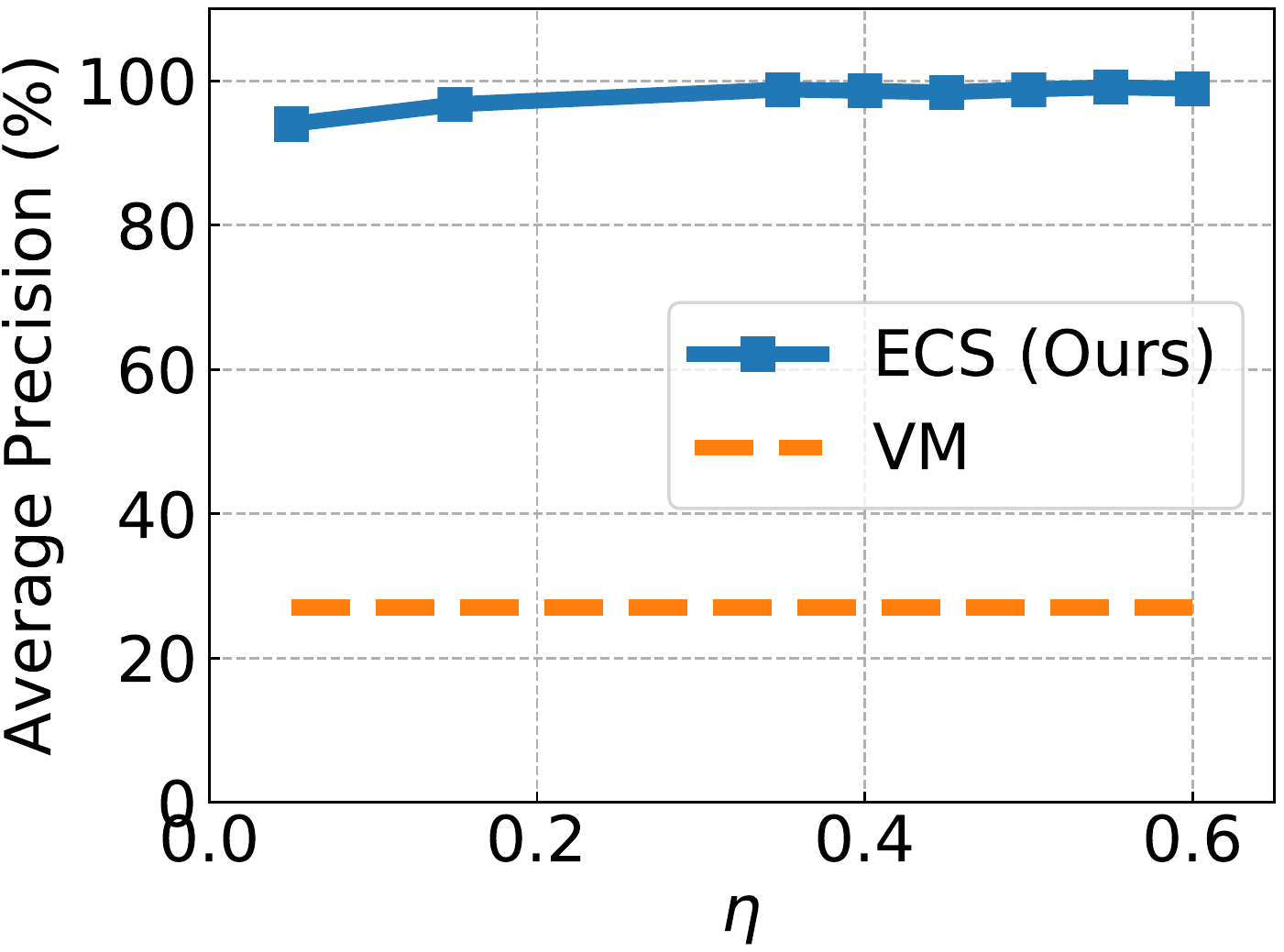}
\includegraphics[width=0.32\columnwidth]{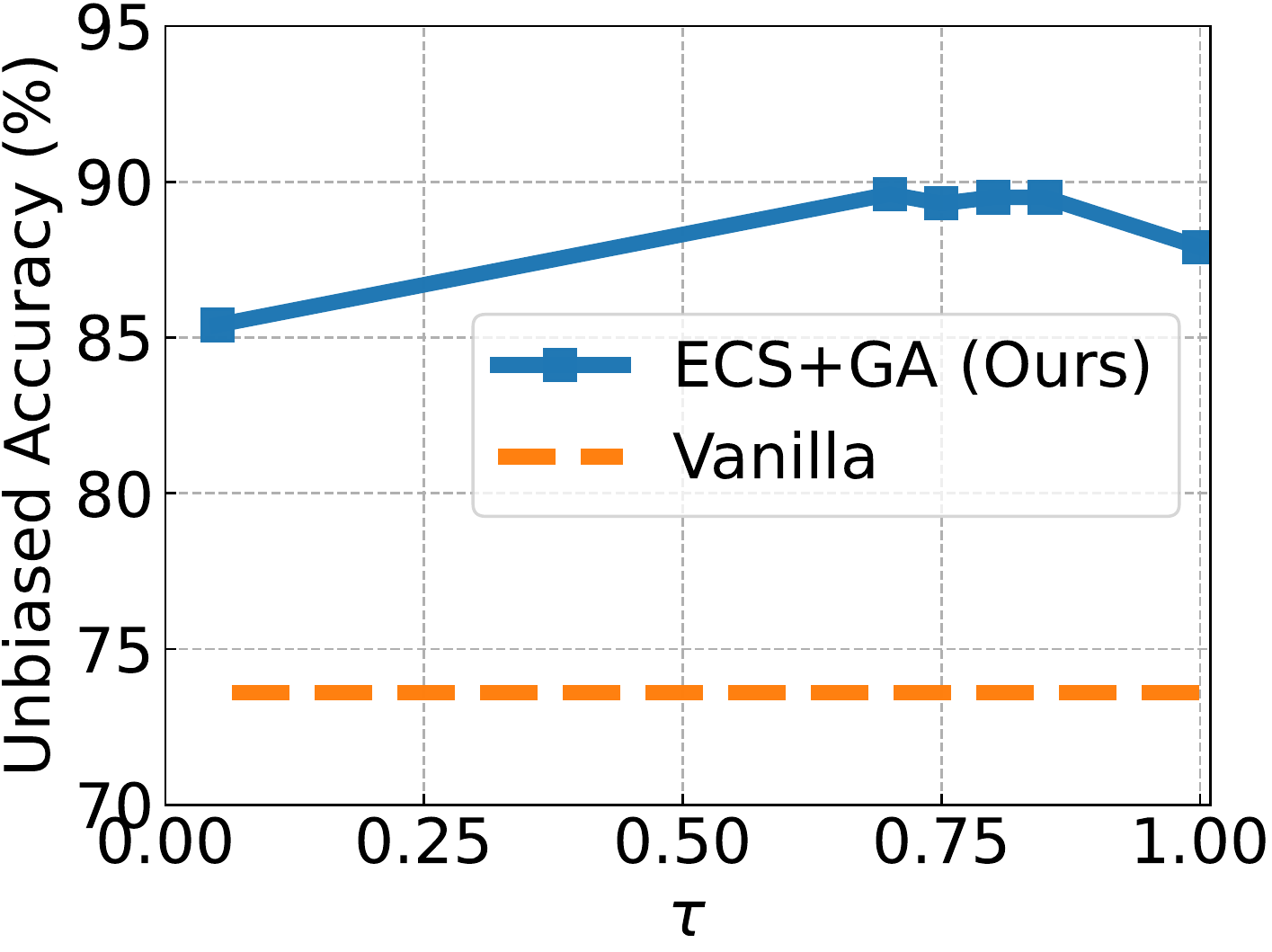}
\includegraphics[width=0.32\columnwidth]{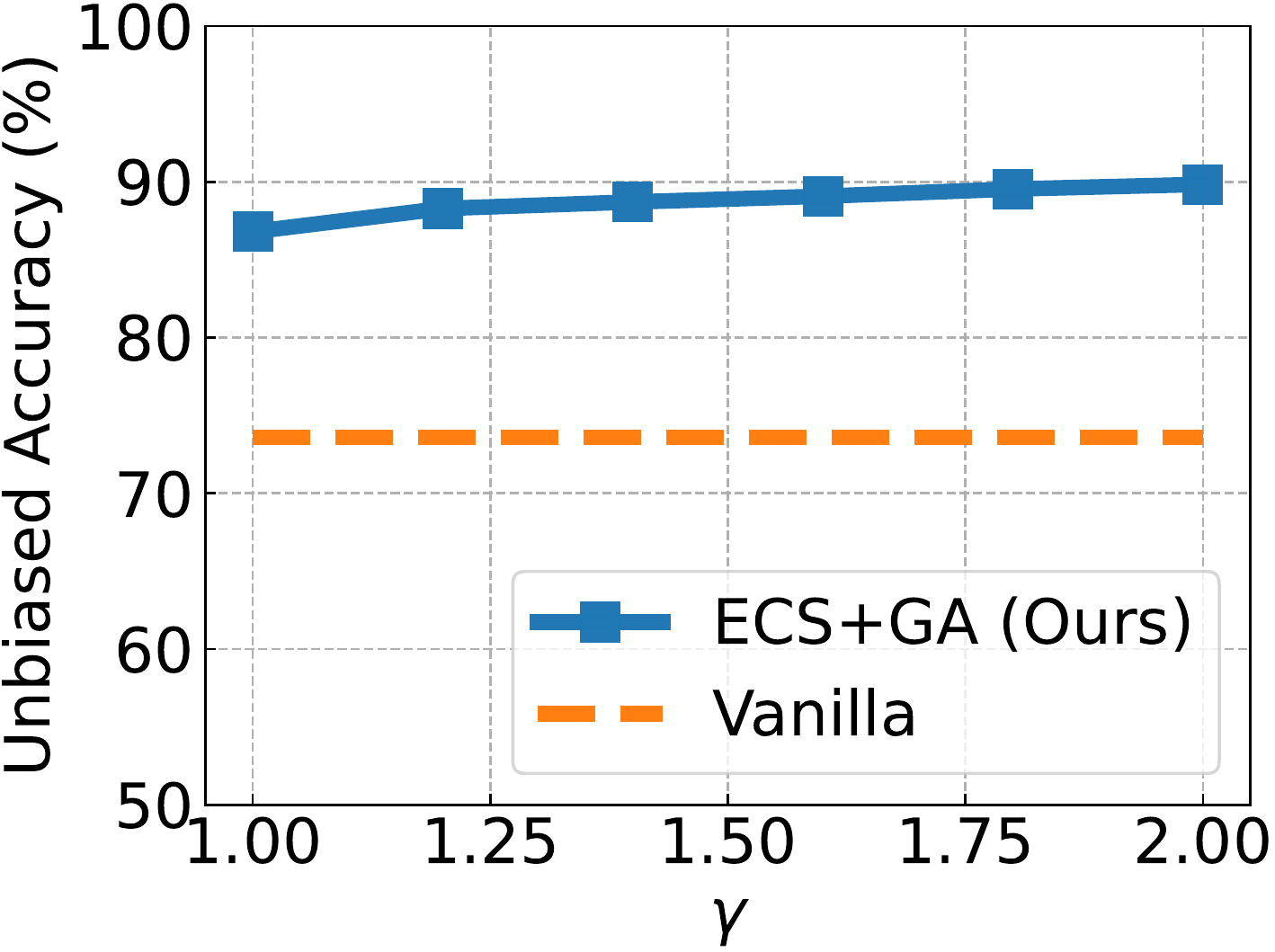}
\caption{Ablation on thresholds $\eta$, $\tau$ and balance ratio $\gamma$.}
\label{fig:vary_param}
\end{figure}

\noindent \textbf{The introduced hyper-parameters are principled and insensitive.} Though the hyper-parameters are critical for methods aimed at combating unknown biases, recent studies~\cite{nam2020learning,kim2021learning} did not include an analysis for them. Here, we present the ablation studies on C-MNIST ($\rho$=98\%) for the hyper-parameters ($\eta$, $\tau$, $\gamma$) in our method as shown in Fig.~\ref{fig:vary_param}. We find that the method performs well under a wide range of hyper-parameters. Specifically, for the confidence threshold $\eta$ in ECS, when $\eta$ $\rightarrow$ 0, most samples will be used to train the auxiliary biased models, including the bias-conflicting ones, resulting in low b-c scores for bias-conflicting samples (\textit{i.e.}, low recall of the mined bias-conflicting samples); when $\eta$ $\rightarrow$ 1, most samples will be discarded, including the relative hard but bias-aligned ones, leading to high b-c scores for bias-aligned samples (\textit{i.e.}, low precision). The determination of $\eta$ is related to the number of categories and the difficulty of tasks, \textit{e.g.}, 0.5 for C-MNIST, 0.1 for C-CIFAR10$^1$ and C-CIFAR10$^2$ (10-class classification tasks), 0.9 for B-Birds and B-CelebA (2-class) here. As depicted in Fig.~\ref{fig:vary_param}, ECS achieves consistent strong mining performance around the empirical value of $\eta$. We also investigate ECS+GA with varying $\tau$. High precision of the mined bias-conflicting samples guarantees that GA can work in stage \uppercase\expandafter{\romannumeral2}, and high recall further increases the diversity of the emphasized samples. Thus, to ensure the precision first, $\tau$ is typically set to 0.8 for all experiments. From Fig.~\ref{fig:vary_param}, ECS+GA is insensitive to $\tau$, however, a too high or too low value can cause low recall or low precision, resulting in inferior performance finally. For the balance ratio $\gamma$, though the results are reported with $\gamma$ = 1.6 for all settings on C-MNIST, C-CIFAR10$^1$ and C-CIFAR10$^2$, 1.0 for B-Birds and B-CelebA, the proposed method is not sensitive to $\gamma$ $\in$ $[1.0, 2.0]$, which is reasonable as $\gamma$ in such region makes the contributions from bias-conflicting samples close to that from bias-aligned samples.

\begin{figure}[t]
\centering
\includegraphics[width=0.7\columnwidth]{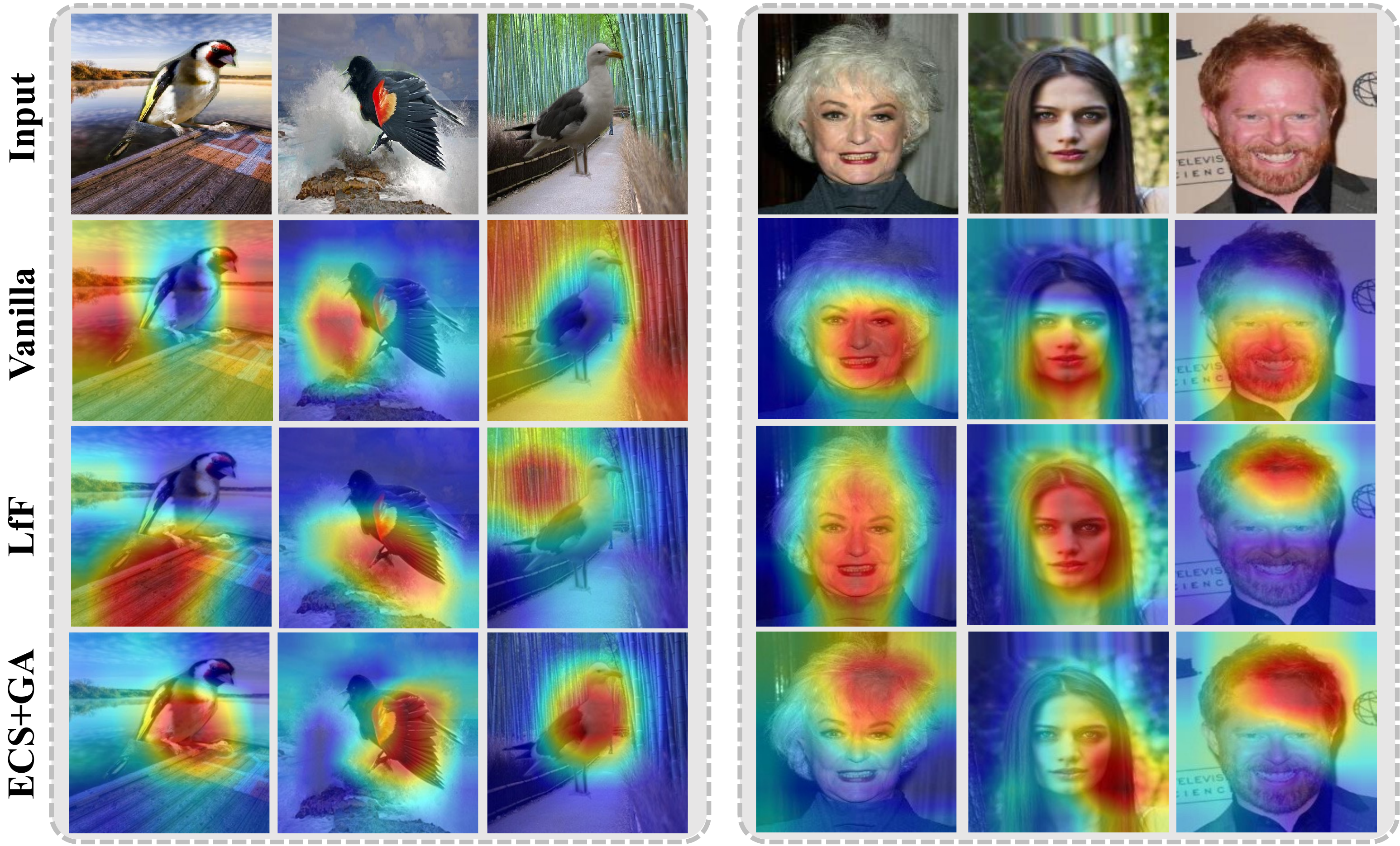}
\caption{Activation maps on Biased Waterbirds (left) and CelebA (right).}
\label{fig:cam}
\end{figure}

\noindent \textbf{The proposed method makes decisions based on intrinsic features.} We visualize the activation maps via CAM~\cite{zhou2016learning} in Fig.~\ref{fig:cam}. Vanilla models usually activate regions related to biases when making predictions, \textit{e.g.}, the background in B-Birds, the gender characteristics (faces, beards) in B-CelebA. LfF can focus attention on key areas in some situations, but there are still some deviations. Meanwhile, the proposed ECS+GA mostly utilizes compact essential features to make decisions.

\section{Conclusions}
\label{sec:dis}
Biased models can cause poor out-of-distribution performance and even negative social impacts. In this paper, we focus on combating unknown biases which is urgently required in realistic applications, and propose an enhanced two-stage debiasing method. In the first stage, an effective bias-conflicting scoring approach containing peer-picking and epoch-ensemble is proposed. Then we derive a new learning objective with the idea of gradient alignment in the second stage, which dynamically balances the gradient contributions from the mined bias-conflicting and bias-aligned samples throughout the learning process. Extensive experiments on synthetic and real-world datasets reveal that the proposed solution outperforms previous methods.

\section*{Acknowledgments}
This work is supported in part by the National Natural Science Foundation of China under Grant 62171248, the R\&D Program of Shenzhen under Grant JCYJ20220818101012025, and the PCNL KEY project (PCL2021A07).

\begin{small}
\bibliography{aaai23}
\end{small}


\end{document}